\documentclass{article}

\usepackage{arxiv}

\usepackage[utf8]{inputenc} % allow utf-8 input
\usepackage[T1]{fontenc}    % use 8-bit T1 fonts
\usepackage{hyperref}       % hyperlinks
\usepackage{url}            % simple URL typesetting
\usepackage{booktabs}       % professional-quality tables
\usepackage{amsfonts}       % blackboard math symbols
\usepackage{nicefrac}       % compact symbols for 1/2, etc.
\usepackage{microtype}      % microtypography
\usepackage{lipsum}
\usepackage{graphicx}
\graphicspath{ {./images/} }

\usepackage{ragged2e}
\usepackage{booktabs,makecell,multirow,tabularx}
\usepackage{arydshln} % 加载虚线支持的宏包
\usepackage{relsize}

\usepackage{amsmath,amssymb,amsfonts}
\usepackage{algorithmic}
\usepackage{graphicx}
\usepackage{textcomp}
\usepackage{subcaption}

\usepackage{hyperref} % 引入 hyperref 宏包
\usepackage{xcolor}   % 引入 xcolor 宏包以自定义颜色

\hypersetup{
	colorlinks=true,   % 启用链接颜色
	linkcolor=purple,    % 内部链接颜色
	urlcolor=cyan,     % URL 链接颜色
	citecolor=green    % 引用链接颜色
}

\title{Prototype Contrastive Consistency Learning for Semi-Supervised Medical Image Segmentation}

\author{
 Shihuan He\textsuperscript{1}, Zhihui Lai\textsuperscript{2}, Ruxin Wang\textsuperscript{1}, Heng Kong\textsuperscript{3} \\ \\
 \textsuperscript{1}Shenzhen Institutes of Advanced Technology, Chinese Academy of Sciences \\
 \textsuperscript{2}Computer Vision Institute, College of Computer Science and Software Engineering, Shenzhen University \\
 \textsuperscript{3}Department of Breast and Thyroid Surgery, Baoan Central Hospital of Shenzhen \\
  \texttt{sh.he3@siat.ac.cn}, \texttt{lai{$\_$}zhi{$\_$}hui@163.com}, \texttt{rx.wang@siat.ac.cn}, \texttt{generaldoc@126.com} 
}

\begin{document}
\maketitle
\begin{abstract}
Medical image segmentation is a crucial task in medical image analysis, but it can be very challenging especially when there are less labeled data but with large unlabeled data. Contrastive learning has proven to be effective for medical image segmentation in semi-supervised learning by constructing contrastive samples from partial pixels.  However, although previous contrastive learning methods can mine semantic information from partial pixels within images, they ignore the whole context information of unlabeled images, which is very important to precise segmentation. In order to solve this problem, we propose a novel prototype contrastive learning method called Prototype Contrastive Consistency Segmentation (PCCS) for semi-supervised medical image segmentation. The core idea is to enforce the prototypes of the same semantic class to be closer and push the prototypes in different semantic classes far away from each other. Specifically, we construct a signed distance map and an uncertainty map from unlabeled images. The signed distance map is used to construct prototypes for contrastive learning, and then we estimate the prototype uncertainty from the uncertainty map as trade-off among prototypes. In order to obtain better prototypes, based on the student-teacher architecture, a new mechanism named prototype updating prototype is designed to assist in updating the prototypes for contrastive learning. In addition, we propose an uncertainty-consistency loss to mine more reliable information from unlabeled data. Extensive experiments on medical image segmentation demonstrate that PCCS achieves better segmentation performance than the state-of-the-art methods. The code is available at \url{https://github.com/comphsh/PCCS}.
\end{abstract}

% keywords can be removed
%\keywords{First keyword \and Second keyword \and More}

\section{Introduction}
In recent years, supervised medical image segmentation has achieved great progress. For example, Fully Convolutional Networks \cite{FCN} is a milestone method in image segmentation. It replaces fully connected layer with convolutional layer to dense prediction for the first time, enabling end-to-end segmentation. 
U-Net \cite{UNet} extracts multi-scale features based on the encoder-decoder structure for medical image segmentation. Inspired by these representative methods, some variants of U-Net \cite{3DUNet,VNet,UNet++} and new architectures \cite{Deeplabv3,HRNet} achieve state-of-the-art performance. These excellent supervised segmentation techniques rely on a large number of annotated images, which are not easy to obtain due to the limitation of professional clinical knowledge and the expensive cost. In contrast, large-scale unlabeled medical images are easy to obtain. Therefore, some studies focus on how to use unlabeled images for segmentation.

Semi-supervised learning methods use a few labeled images and a large number of unlabeled images to improve the performance of the model and obtain impressive results in image classification and semantic segmentation. The learning strategies of semi-supervised image segmentation include co-training \cite{Co-train-DCT-segmentation} using different views, consistency regularization \cite{MCNet-consistency} using slight perturbations, pseudo-label \cite{pseudo-CPS} training using prediction as supervised information, entropy minimization \cite{EM} and so on. However, these methods only pay attention to the same position relationship between images and ignore the information among pixels between the labeled and unlabeled images.

In order to explore the information between image pixels, contrastive learning has been proposed for image segmentation, which aims to improve the representation ability of the encoder so that similar objects in the embedding space are close to each other and dissimilar objects are far away from each other. The key strategy in contrastive learning is how to define similar samples and dissimilar samples. In image-level classification tasks, MoCo \cite{contra-Moco1} and SimCLR \cite{contra-SimCLR} get different views through different transformations. However, semantic segmentation requires classifying each pixel to the corresponding label. To address this problem, pixel-level contrastive learning is used to improve segmentation accuracy. Global and local features of the labeled images are used in contrastive learning \cite{contra--gloab-locall}. ContraSeg \cite{contra-explore-cross} proposes to explore the cross-image pixels to construct the contrastive samples for contrastive learning with pixel-level sampling from different labeled images. After that, \cite{contra-PC2Seg} and ~\cite{contra-label-effient} introduce contrastive learning into semi-supervised segmentation and use the pseudo label to construct positive and negative samples so that contrastive learning can also utilize unlabeled data. To take advantage of high-quality pseudo-labels, \cite{UGPCL} utilizes uncertainty maps to guide the construction of contrastive samples, while \cite{pseudo-contra-U2PL} improves pseudo-labels' quality by mining the information of unreliable labels. \cite{Tripled_Uncertainty} uses the average of multiple uncertainty values to generate high-quality pseudo labels for contrastive learning.

\begin{figure}[t]
	\centerline{\includegraphics[width=4.5in]{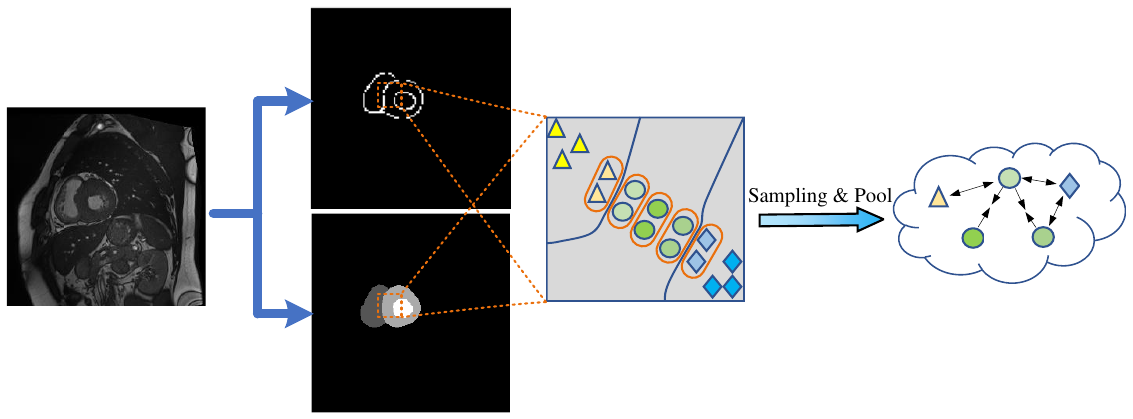}}
	\caption{Workflow of prototype contrastive learning. We aim to enhance the contrastive ability of prototypes to improve the effectiveness of contrastive learning.}
	\label{fig:contra_detail}
\end{figure}

However, since the predicted pseudo-label may be inconsistent with the ground-truth semantic label, pixel contrastive learning in semi-supervised case is sensitive to noise from mispredicted pixels, leading to error accumulation. In addition, the memory and computational cost of pixel-level contrastive learning are {$\mathcal{O}$}({$N^2$})({$N$} is the image size), which usually requires complex filtering mechanisms in sampling negative and positive samples. \cite{contra-rethink-prototype} designs contrastive samples from the prototype of the pixels in the labeled image, which is only suitable for supervised learning. \cite{contra-semi-proto} and \cite{SCPNet} introduce prototypes into contrastive learning, and use the predictions of prototypes as pseudo-labels for contrastive learning. However, these methods simply view the pixels in the same class as the prototype without fully mining the internal information of the image. Therefore, discriminative information among boundary prototypes has not been explored in previous semi-supervised prototype contrastive learning to a large extent, and how to design discriminative boundary prototypes to improve medical image segmentation is still an open problem.

To overcome the above challenges, as shown in Fig.~\ref{fig:contra_detail}, our core idea is to build relationships among boundary prototypes from the feature map and improve the discriminative ability of prototypes using unlabeled data and labeled data for contrastive learning. Thus, we propose a novel semi-supervised medical image segmentation method named Prototype Contrastive Consistency Segmentation (PCCS). The proposed PCCS uses different prototypes through signed distance maps and pseudo-labels for contrastive learning. To reduce error accumulation caused by pseudo-label, we assign weights to each prototype using the uncertainty map. For obtaining better prototypes, a new prototype-updating mechanism named prototype updating prototype is designed to update the prototype layer with the feature map from the encoder. To enhance the quality of generated pseudo-labels, we use constrained consistency learning to drive the model to obtain better predictions of medical image segmentation in the training stage.

In summary, our contributions mainly include:
\begin{itemize}
	\item By making full use of contrastive consistency of unlabeled data, a novel semi-supervised medical image segmentation called PCCS is proposed, which can significantly improve the prototype's discriminative ability so as to make intra-class features more compact and inter-class features more separable. 
	\item A new mechanism for updating prototype based on student-teacher architecture is designed, which can guide the model to enhance the diversity of prototype from unlabeled data so as to promote the representation ability of the encoder for accurate image segmentation.
	\item An uncertainty-consistency loss is designed to reduce the uncertainty of prediction so as to generate precise segmentation mask. Experimental results show that the proposed PCCS can obtain competitive performance against the state-of-the-art methods on some medical image segmentation datasets.
\end{itemize}

\section{Related Work}\label{sec:related work}
\subsection{Contrastive Learning}
Self-supervised contrastive learning has achieved great success in image segmentation.  In contrastive learning, how to construct contrastive samples is the key operation. There are various strategies to select appropriate contrastive samples in existing work. The early sampling strategies \cite{contra-Moco1,contra-Moco2} use a memory bank and momentum encoder to provide negative samples. Chen \cite{contra-SimCLR} select contrastive samples from multiple batches. However, since each pixel in semantic segmentation needs to be classified, the references in \cite{contra-explore-cross,contra-label-effient,contra-PC2Seg,contra--gloab-locall,contra-rethink-prototype} extend image-level contrastive learning to pixel-level contrastive learning for semantic segmentation. Another representative strategy is to use the label of annotated images to determine the contrastive pixels \cite{contra-explore-cross}. 
To further improve the performance, references \cite{contra-label-effient,contra-PC2Seg,contra-region-level,LocalCL} explore pixel contrastive learning using pseudo-label on semi-supervised semantic segmentation. Unreliable pseudo-labels are used to sample negative pixels for contrastive learning in \cite{pseudo-contra-U2PL}. These semi-supervised methods focus on sampling partial pixels and usually need to design complex sampling mechanisms in selecting contrastive samples. Chen and Lian \cite{contra-semi-proto} simply view the same class's pixels as prototypes, which ignores semantic information of intra-prototypes. To alleviate this problem, a feasible strategy is to construct contrastive samples using signed distance map of the model's prediction and generate discriminative prototypes by fully utilizing all pixels of the image in contrastive learning.

\begin{figure*}[htbp]
	\centerline{\includegraphics[width=6.4in]{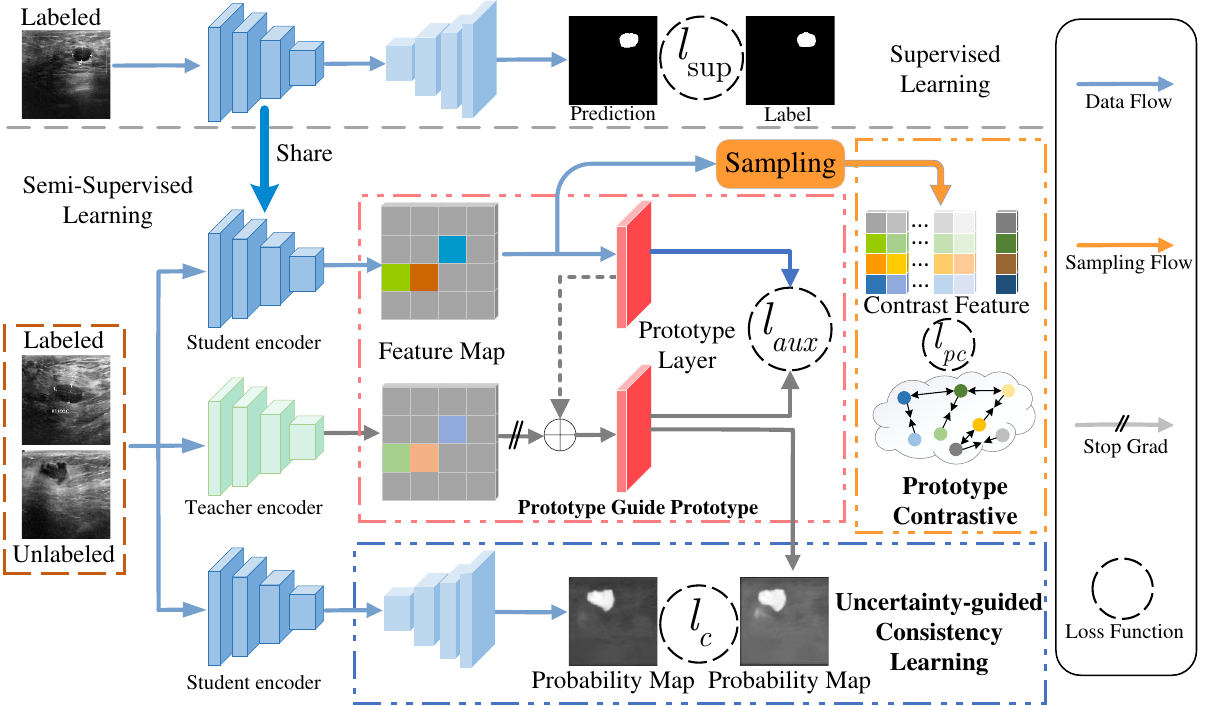}}
	\caption{Overview of the PCCS. Arrows of different colors indicate the processing flow of different data. PCCS includes three modules, they are prototype contrastive learning module, prototype guide prototype module, and uncertainty-guided consistency learning module. The prototype contrastive learning module obtains contrastive samples from the feature map of the encoder and performs uncertainty-weighted prototype contrastive consistency loss {$l_{pc}$}. The prototype guide prototype module can enhance the diversity of prototype and improve generalization ability by aux loss {$l_{aux}$}. 
		Uncertainty-guided consistency learning module enforces the model to make a consistent prediction for the output of two branches and reduce the uncertainty of the prediction by  uncertainty-consistency loss {$l_{c}$}.}
	\label{fig:main_arch}
\end{figure*}

\subsection{Semi-supervised medical image segmentation}
To reduce the burden of manual annotation, many semi-supervised medical image segmentation methods have been proposed, which use a few labeled data and a large amount of unlabeled data. Existing semi-supervised methods mainly include pseudo-label learning \cite{pseudo-CPS}, collaborative training \cite{Co-train-DCT-segmentation,Co-train-DCT-recognition} and consistency learning \cite{Consis-MT,Consis-Uncertainty-UA-MT,Consis-Transform-consis}, entropy minimization methods \cite{EM}. Pseudo-label methods view predictions of the model for unlabeled images as pseudo-labels and then use the predicted pseudo-labels to supervise the model. 
Reference \cite{pseudo-CPS} proposes a strategy that pseudo-labels and network parameters are updated iteratively. Adversarial learning is introduced to medical image segmentation in \cite{Adversarial-Net}, encouraging the segmentation output of unlabeled data to be similar to the labeled data. Entropy minimization method \cite{EM} believes that high-quality prediction results should have smaller entropy, so it performs model learning by minimizing the information entropy of the predicted probability distribution. Co-training \cite{Co-train-DCT-segmentation} assumes that multiple decision views contain complementary information so that different classifiers are designed to learn different views to improve segmentation performance.

\subsection{Consistency Learning}
The effectiveness of consistency learning is based on the assumption that data points close to each other are likely to be from the same class. Based on this assumption, the output results of the model should make consistent predictions even if the input images have various perturbations, such as input perturbations, model perturbations, and so on.
The references in \cite{Consis-MT,Consis-Temporal,Consis-Uncertainty-UA-MT} construct different perturbed images through data augmentation for consistency learning. Luo \cite{Uncertainty-pyramid-consis}  obtains perturbed versions at different feature scales from the decoder. Some others \cite{bound-dual-task,bound-shape-aware}  perform consistency learning across multiple decoders in different tasks. The above methods make great progress in semantic segmentation. However, they do not take the prototype's contrastive consistency into consideration, leading to degrading segmentation performance.

\section{Method}
Since the framework is used for semi-supervised medical image segmentation, we first give an overview description of the PCCS as shown in Fig.~\ref{fig:main_arch}. Then, we will introduce Prototype Contrastive Learning, Prototype Guide Prototype, and Uncertainty-guided Consistency Learning. Prototype Contrastive Learning pulls prototypes in the same class closer and pushes the prototypes in different classes far away from each other. At the same time, Prototype Guide Prototype enhances the diversity of prototypes and improves the representation ability of prototypes.  Finally, Uncertainty-guided Consistency Learning will enforce the consistency of predictions and reduce the uncertainty of predictions.

\subsection{Overview of the PCCS Architecture}
The overall architecture of our prototype contrastive consistency semi-supervised medical image segmentation method is shown in Fig.~\ref{fig:main_arch}. Let {$\mathcal{D}_{l}=\left\{\boldsymbol{x}_{i}, \boldsymbol{y}_{i}\right\} _{i=1}^{n_{l}}$}  and {$\mathcal{D}_{u}=\left\{\boldsymbol{x}_{j}\right\}_{ j=1}^{n_{u}}$} denote labeled dataset and unlabeled data respectively, and {$\boldsymbol{x}_i,\boldsymbol{x}_j \in \mathbb{R}^{H \times W}$} represent the training images of size {$H \times W$},
{$\boldsymbol{y}_{i} \in \{0,1\}^{H \times W \times C}$} represents the {$C$}-th class pixel-level label corresponding to image {$\boldsymbol{x}_{i}$}. The segmentation network {$h$} is composed of the encoder {$f$} and the decoder {$g$} such that {$h=g \odot f$}. Therefore, {$\hat{\boldsymbol{y}_i}=h\left(\boldsymbol{x}_{i}\right)$} is the model's prediction for image {$\boldsymbol{x}_{i}$}.
The proposed method is divided into two steps. In the first step, the model is trained using labeled images by minimizing the following  cross-entropy loss and dice loss:
\begin{equation}
	l_{sup}=\frac{1}{2}(L_{CE}\left(\boldsymbol{y}_{i},\hat{\boldsymbol{y}_i}\right)+ L_{Dice}\left(\boldsymbol{y}_{i},\hat{\boldsymbol{y}_i}\right)),
	\label{eq:sup_loss}
\end{equation}
where {$L_{CE}$} is the cross-entropy loss and {$L_{Dice}$} is the dice loss. In the second step, the model is trained using unlabeled images and labeled images based on the learned network parameters in the first step. In this paper, we elaborately design a contrastive consistency model to exploit unlabeled images, which includes three modules, i.e. prototype contrastive learning module, prototype guide prototype module, and uncertainty-guided consistency learning module.

For the prototype contrastive learning module, we propose a novel sampling strategy to obtain contrastive samples and use uncertainty-weighted prototype contrastive consistency loss {$l_{pc}$} to make positive samples close to each other and negative samples far from each other.
For the prototype guide prototype module, a new mechanism is designed to update the prototypes with the features of the encoder so as to enhance the diversity of the prototypes and improve generalization ability. 
For the uncertainty-guided consistency learning module, unlabeled images are augmented to obtain an augmented image. The original image is fed to the student branch, and the augmented image is fed to the teacher branch. The uncertainty-consistency loss {$l_{c}$} enforces the model to make a consistent prediction for the output of two branches and reduce the uncertainty of the prediction. In summary, the overall objective function of the proposed method is:
\begin{equation}
	l_{total} = l_{sup} + \lambda_cl_{c} + \lambda_{aux}l_{aux}  + \lambda_{pc} l_{pc},
	\label{eq:tot_loss}
\end{equation}
where {$\lambda_{c}$}, {$\lambda_{aux}$} and {$\lambda_{pc}$} are trade-off hyperparameters. According to the experiment, we usually set {$\lambda_{aux}=0.3$} and {$\lambda_{pc}=0.1$}. Specifically, {$\lambda_{c} = 0.1e^{-5(1-t/30000)^{2}}$}, where {$t$} represents the current iteration number.

\subsection{Prototype Contrastive Learning}
The images are fed into the student network to extract multi-scale features map {$\mathcal{M} \in \mathbb{R}^{H \times W \times D}$}, where {$H \times W$} is the size of the input image, DD is the feature dimension. Since the feature dimension {$D$} of the feature map generated by the encoder is usually large, in order to reduce the memory cost,  the feature map {$\mathcal{M}$} goes through a projection layer, which reduces the feature dimension of {$\mathcal{M}$} from DD to 128, that is {$\mathcal{M}^{f} \in \mathbb{R}^{H \times W \times 128}$}. The proposed module selects pixel features from {$\mathcal{M}^{f}$},  which is used in the proposed anchor sampling strategy and prototype contrastive consistency loss. The details are introduced as below.

\begin{figure*}[htbp]
	\centerline{\includegraphics[width=6.4in]{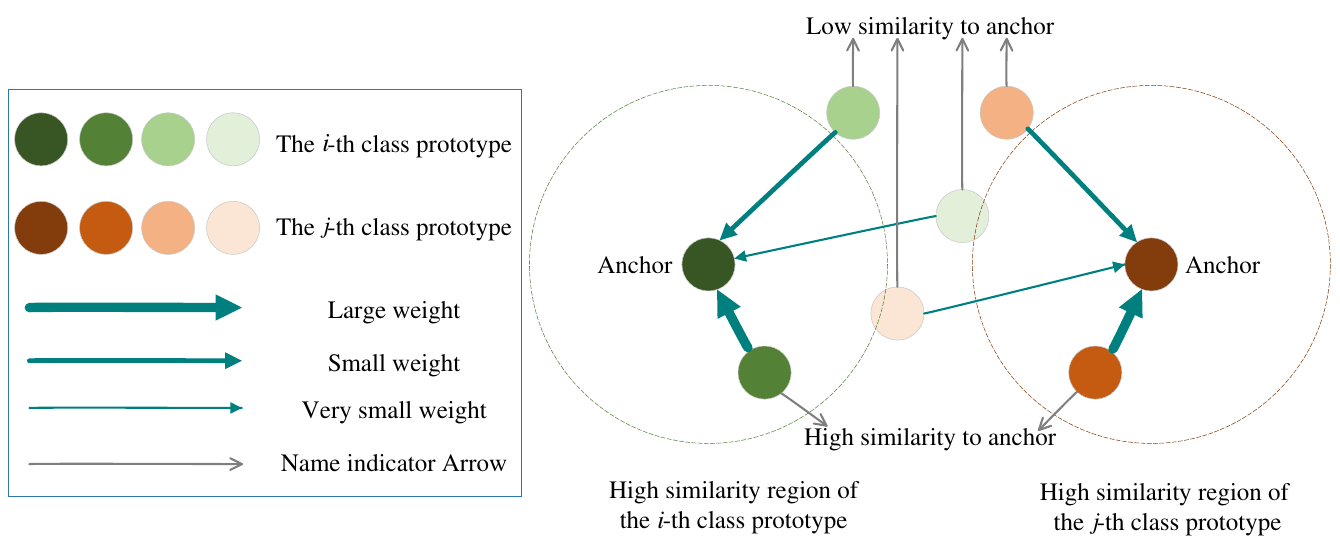}}
	\caption{Overview of prototype contrastive consistency loss. Large weights are assigned to positive prototypes with high similarity to the anchor, while small weights are assigned to positive prototypes with low similarity to the anchor. In this way, the positive samples with high similarity are kept close to the anchor and the positive samples with low similarity are suppressed, which can alleviate the impact of wrong pseudo-labels and learn a good feature space structure to achieve intra-class compactness and inter-class separability.}
	\label{fig:prototype_contrast_consistency}
\end{figure*}

{\raggedleft \bf Anchor Sampling in boundary pixels.} The purpose of prototype contrastive learning is to enhance the similarity between similar semantic prototypes in images and to push prototypes of different semantic categories far away from each other. Therefore, the key idea is how to define the prototype. In PCCS, the prototypes are defined as the mean feature of all pixels in the embedding space with equal distances from the boundary. The distance between each position (pixel) and the boundary of segmentation within the image is calculated by
\begin{equation}
	\mathcal{T}(y)=\left\{
	\begin{array}{lr}
		-\underset{b \in \partial S}{inf}\|a-b\|_{2}, & a \in S_{i n} \\
		0, & a \in \partial S \\
		+\underset{b \in \partial S}{i n f}\|a-b\|_{2}, & a \in S_{out}
	\end{array}\right.,
	\label{eq:T(x)}
\end{equation}
where {$y$} is the segmentation mask. {$a$}, {$b$} are two different pixels in a segmentation mask {$y$}. {$\partial S$} is the boundary in the segmentation mask and represents the silhouette of the object, {$S_{in}$} and {$S_{out}$} denote the target object region and other regions respectively. {$\mathcal{T}(y)$} is the transformation from a segmentation mask to a signed distance map. Let {$\mathcal{S}_{c} \in  \mathbb{R}^{H \times W}$} denote the signed distance map of the {$c$}-th class in the segmentation mask, {$\mathcal{S}_{c,i}$} represents the value of the signed distance map {$\mathcal{S}_{c}$} at index {$i$}, where {$i \in \{1,2,\cdots,H \times W\}$}. Let {$\mathcal{N}_{c, j}$} denote the number of pixels of the {$c$}-th class with distance {$j$} from the boundary, which can be calculated by
\begin{equation}
	\mathcal{N}_{c, j}=\sum_{i=1}^{H \times W} \boldsymbol{1}\left\{\mathcal{S}_{c, i}=j\right\},
	\label{eq:N_c}
\end{equation}
where {$\boldsymbol{1}\left\{ \cdot \right\}$} is an indicator function that determines whether an expression is true or false and returns 1 or 0. If {$\mathcal{S}_{c, i}=j$} is true, then {$\boldsymbol{1}\left\{\mathcal{S}_{c, i}=j\right\}$} is 1. If {$\mathcal{S}_{c, i}=j$} is false, {$\boldsymbol{1}\left\{\mathcal{S}_{c, i}=j\right\}$} is 0. The number of pixels whose distance to the segmentation boundary is {$j$} in the signed distance map {$\mathcal{S}_{c}$} can be calculated by Equation \ref{eq:N_c}. The boundary prototype is the mean feature of sample features with the same semantics, which is computed by
\begin{equation}
	{p}_{c, j}=\frac{1}{\mathcal{N}_{c, j}} \sum_{i=1}^{H \times W} \boldsymbol{1}\left\{\mathcal{S}_{c, i}=j\right\} \cdot \mathcal{M}_{i}^{f},
	\label{eq:p_cj}
\end{equation}
where {$\mathcal{M}_{i}^{f}$} is a feature at index {$i$} of the feature map {$\mathcal{M}^{f}$}, prototype {${p}_{c, j}$} is the mean feature of all the pixel features of the {$c$}-th class with the same distance {$j$} in the signed distance map. All prototypes calculated from Equation \ref{eq:p_cj} form the set {$\mathcal{P}=\left\{p_{c, j} \mid c \in\{1,2, \cdots, C\}, j \in \mathcal{D}_{c}\right\}$}, where the set {$\mathcal{D}_{c}$} contains all values of the signed distance map on the {$c$}-th class and it can be denoted as
\begin{equation}
	\mathcal{D}_{c}=\left\{\mathcal{S}_{c, i} \mid i \in\{1,2, \cdots, H \times W\}, \mathcal{S}_{c, i}<0\right\}.
	\label{eq:D_c}
\end{equation}

In PCCS, the anchors used for contrastive learning are obtained from {$\mathcal{P}$}. In other words, each prototype in {$\mathcal{P}$} acts as an anchor. For any prototype {${p}_{c, j}$} in {$\mathcal{P}$}, its positive samples construct the positive sample set {$\mathcal{P}^{+}=\left\{p_{c, u} \mid u \in \mathcal{D}_{c}\right\}-\left\{p_{c, j}\right\}$}, the negative samples of prototype {${p}_{c, j}$} are taken from the negative sample set {$\mathcal{P}^{-}=\mathcal{P}-\mathcal{P}_{c, j}^{+}$}.

{\raggedleft \bf Uncertainty-weighted Prototype Contrastive Consistency Loss.} The key operation of prototype contrastive learning is to keep the distance between anchor prototype {${p}_{c, j}$} and the positive prototype from {$\mathcal{P}^{+}$} to be closer, and and push anchor prototype {${p}_{c, j}$} away from the negative prototype from {$\mathcal{P}^{-}$}.
However, the computation of the prototype depends on the pseudo-label assigned to the unlabeled image, which may not always be entirely accurate. Incorrect pseudo-label can lead to the prototype being erroneously closer to the negative sample and farther from the positive sample, thereby hindering the effective achievement of intra-class compactness and inter-class separability. As shown in Fig.~\ref{fig:prototype_contrast_consistency}, we consistently aim for the anchor prototype to maintain proximity to positive samples, distance itself from negative samples, and remain unaffected by erroneous pseudo-labels. To achieve this goal, we introduce an uncertainty-weighted prototype contrastive consistency loss, including two components: contrastive consistency loss and uncertainty weighting. The contrastive consistency loss is defined as follows:
\begin{equation}
	\mathcal{L}_{pc}(p_{c,j})\!=-\sum_{ p^+  \in \mathcal{P}^+}
	{\mathcal{S}(p^+)}
	\log\frac{e^{ p_{c,j}   p^{+}  / \tau}}
	{e^{ p_{c,j}   p^{+}  / \tau}+ \sum_{ p^{-} \in \mathcal{P}^-}e^{p_{c,j}  p^{-}/ \tau}}
	\label{eq:L_pc}
\end{equation}
where 
\begin{equation}
	\mathcal{S}(p^+)=\frac{p_{c,j}  p^{+}}{\sum_{ p^+ \in \mathcal{P}^+}{p_{c,j}  p^{+}}},
	\label{eq:S()}
\end{equation}
where {$ p_{c,j} $}, {$p^{+} $}, and {$p^{-}$} used to calculate {$\mathcal{L}_{pc}(p_{c,j})$} are normalized to be {$[0,1]$}, {$\tau$} is a temperature hyperparameter that controls the probability distribution of the vector. 
{$p_{c,j} p^{+}$} represents the similarity between the anchor prototype {$p_{c,j}$} and the positive sample prototype {$p^{+}$}. {$\mathcal{S}(p^+)$} denotes the proportion of the positive sample {$p^{+}$} in the total similarity {${\sum_{ p^+ \in \mathcal{P}^+}{p_{c,j} p^{+}}}$}.
If the similarity between the positive sample and the anchor prototype is greater, then {$\mathcal{S}(p^+)$} is larger. Consequently, the contrastive consistency loss pulls the positive sample closer to the anchor prototype. Conversely, if the similarity is smaller, the impact of the positive sample with lower similarity on the contrastive consistency loss is reduced.

In order to reduce the impact of pseudo-labels on the prototype, we further design an uncertainty-weighted contrastive consistency loss, which is defined as:
\begin{equation}
	l_{pc}=\sum_{p_{c,j}\in \mathcal{P}} \frac{e^{-\mathcal{H}(p_{c,j})}}{\sum_{p \in \mathcal{P}}{e^{-\mathcal{H}(p)}}}\mathcal{L}_{pc}(p_{c,j}),
	\label{eq:l_pc}
\end{equation}
where
\begin{equation}
	\mathcal{H}(p_{c,j})=-\sum\limits_{i=1}^{C}v_i\log v_i,v_i=g_i(p_{c,j}),
	\label{eq:H()}
\end{equation}
where {$g(\cdot )$} is a probability mapping that maps the prototype to the prediction probability vector in {$v_i=g_i({{p}_{c,j}})\in {{\mathbb{R}} ^{C}}$}, {${{v}_{i}}$} represents the probability value of the predicted probability vector in the {$i$}-th class, and {$C$} is the total number of classes. {$\mathcal{H}({{p}_{c,j}})$} represents the total uncertainty value of prototype {${{p}_{c,j}}$} on {$C$} classes. The total uncertainty value for wrong pseudo-labels is larger and the total uncertainty value for correct labels is smaller. When the total uncertainty of the prototype is large, more incorrect pseudo-labels may be used. By degrading the weight of the uncertainty prototype in the contrastive learning loss, the impact of wrong pseudo-labels on prototype contrastive learning is weakened. On the contrary, when the total uncertainty of a prototype is small, it indicates that the credibility of the prototype is higher. The weight of the prototype in contrastive learning loss is increased to make full use of high-quality pseudo labels. In this way, the model boosts the discriminative ability of prototypes via Equation~\ref{eq:l_pc} and makes the intra-class features more compact and inter-class features more separable in the embedding space. According to the experiments, {$\tau$} can be set as 0.05. 

\begin{figure*}[htbp]
	\centerline{\includegraphics[width=4.5in]{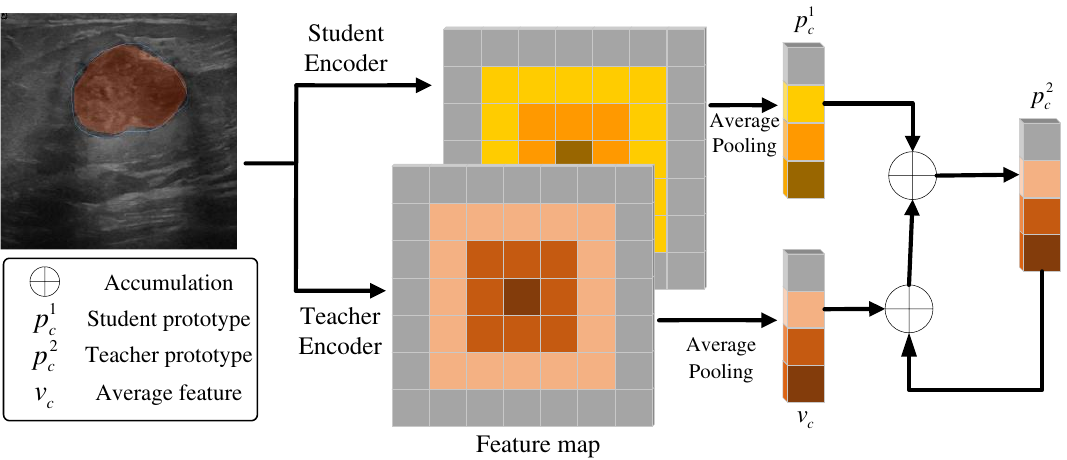}}
	\caption{Overview of the Prototype Guide Prototype Module.}
	\label{fig:prototype_guide}
\end{figure*}

\subsection{Prototype Guide Prototype}
{\raggedleft \bf Prototype-updating Mechanism.} Let {${p}_c^1$} and {${p}_c^2$} denote the student prototype and teacher prototype from different branches of the {$c$}-th class respectively. Inspired by Mean Teacher \cite{Consis-MT}, we can update {${p}_c^2$} for the teacher network branch as follows:
\begin{equation}
	{p}_c^2 = (1-\mu ){p}_c^1  + \mu {p}_c^2,
	\label{eq:alpha2_alpha1}
\end{equation}
where {$\mu$} is the trade-off parameter. Previous research \cite{Consis-Uncertainty-UA-MT} indicates that {$\mu$} can be set as 0.99. However, as the training progresses, {${p}_c^2$} receives significant feedback from the student network, thereby biasing towards the student prototype {${p}_c^1$}, leading to the gradual forgetting of the original information from the teacher and greatly reducing the diversity of prototypes. To increase the diversity of prototypes, We design to integrate the historical information {$v_c$} of the teacher branch and the student prototype {${p}_c^1$} into the teacher prototype {${p}_c^2$}, The overview of the mechanism of updating prototype is shown in Fig.~\ref{fig:prototype_guide} and can be generalized as
\begin{equation}
	{p}_c^2 =  (\mu + \gamma) {p}_c^2 + (1-\mu ){p}_c^1 +(1-\gamma )v_c,
	\label{eq:alpha2_compresentive}
\end{equation}
where we can set {$\mu=0.99$} and {$\gamma=0.999$} according to the experiment. The meaning of Equation \ref{eq:alpha2_compresentive} is that teacher prototype {${p}_c^2$} not only absorbs the student prototype {${p}_c^1$}, but also adjusts the prototype according to the historical information between the features {$v_c$} from the teacher network in the embedding space. This model allows the teacher prototype to maintain greater diversity without collapsing to the trivial solution as the student prototype. This mechanism makes the learned prototype more effective, which is helpful for further boosting the potential of prototype contrastive learning.

{\raggedleft \bf Pixel-Prototype Contrast.} For the training data, let {$v_i$} denote the feature vector of pixel {$i$} mapped by the student encoder.Then, suppose the teacher prototype matching {$v_i$} serves as a positive sample {$p_{c,i}$}, and other prototypes form a negative sample set {$\mathcal{Q}$}. The pixel-prototype loss of a pixel between the student network and the teacher network can be calculated by

\begin{equation}
	\mathcal{L}_{\mathrm{aux}}^{i}=-\log \frac{e^{\cos{( v_i,\ p_{c,i})} / \tau } }{e^{\cos{( v_i , p_{c,i})} / \tau }+\sum_{ q^{-} \in \mathcal{Q}} e^{\cos{( v_i , q^-)} / \tau }},
	\label{eq:L_auxce}
\end{equation}
where the pixel is close to the teacher prototype with high similarity and far away from the teacher prototype with low similarity. In summary, the auxiliary loss of an image $\boldsymbol{x} \in \mathbb{R}^{H \times W}$ can be defined as
\begin{equation}
	l_{aux} = \frac{1}{H \times W}  \sum_{i=1}^{H \times W}  \mathcal{L}_{\mathrm{aux}}^{i}.
	\label{eq:l_aux}
\end{equation}

The student branch's pixels will be closer to the high-certainty teacher prototype and farther away from the low-certainty teacher prototype by the auxiliary loss {$l_{aux}$}, which can effectively mine internal information of unlabeled images.

\subsection{Uncertainty-guided Consistency Learning}
{\raggedleft \bf Output Consistency of Two Networks.} Given an input image for two network branches, we can obtain two predicted probability maps {${p}_{s}$} and {${p}_{t}$} of the student branch and teacher branch, respectively. For the training images, the consistency loss of the two predicted probability maps is calculated by
\begin{equation}
	l_{con}=\mathcal{L}_{\text {dis }}\left(p_{s}, p_{t}\right),
	\label{eq:l_con}
\end{equation}
where {$\mathcal{L}_{\text {dis}}$} represents the mean square error (MSE).

{\raggedleft \bf Minimize the Prediction Uncertainty.} In order to further reduce the prediction uncertainty of the network and mine more reliable information from unlabeled data, an uncertainty-consistency loss is designed to minimize prediction uncertainty.
Let {$v^{i} \in \mathbb{R}^{C}$}
is the predicted probability vector of the {$i$}-th pixel, where {$C$} is the total number of classes. The prediction uncertainty of pixel {$i$} can be denoted as
\begin{equation}
	u^{i}=-\sum_{c} {v}_{c}^{i} \log \left({v}_{c}^{i}+\epsilon\right),
	\label{eq:u_c}
\end{equation}
where {${c} \in \{0,1,...,C-1 \}$} represents the {$c$}-th class. To minimize the uncertainty of the output between the two network branches (student branch and teacher branch), the proposed uncertainty-consistency loss can be summarized as
\begin{equation}
	l_{c}= l_{con} + \lambda _{u}l_{u},
	\label{eq:l_c}
\end{equation}
where 
\begin{equation}
	l_{u}= \frac{1}{2(H \times W)}(\sqrt{\sum_{i}^{H \times W} (u_{s}^{i})^{2}} + \sqrt{\sum_{i}^{H \times W} (u_{t}^{i})^{2}}),
	\label{eq:l_u}
\end{equation}
where {$u_{s}^{i}$} and {$u_{t}^{i}$} are the prediction uncertainty of the student branch and teacher branch at pixel {$i$} within the training image respectively. The mean uncertainty in Equation \ref{eq:l_u} is obtained by averaging the uncertainty of the two network outputs. {$\lambda _{u}$} is the coefficient as a trade-off between uncertainty and consistency loss, which can be set as {$\lambda _{u}=0.01$} in the experiment. In this way, the model can effectively degrade the uncertainty of prediction and further promote the prediction accuracy of pseudo-label so as to yield precise segmentation.

\begin{table}[ht]
	\centering
	% \tiny
	% \scriptsize
	\footnotesize
	% \small
	% \normalsize
	% \large
	
	\caption{Data distribution of the train set, validation set, and test set.}
	\setlength{\tabcolsep}{0.2mm}{
		\begin{tabular}{cccccc}
			\toprule
			Dataset               & Category  & Train & Val & Test & Total \\ \hline
			\multirow{3}{*}{BUSI} & Benign    & 305   & 43  & 89   & 437   \\
			& Malignant & 147   & 21  & 42   & 210   \\
			& Total     & 452   & 64  & 131  & 647   \\ \hline
			\multirow{4}{*}{BML} & MIP       & 351   & 50  & 100  & 501   \\
			& T1a       & 420   & 59  & 118  & 597   \\
			& T2        & 405   & 57  & 114  & 576   \\
			% & Total     & 1176  & 166 & 332  & 1674  \\ \cline{2-6} 
			% & L2        & 242   & 29  & 77   & 348   \\
			% & L3        & 253   & 46  & 76   & 375   \\
			% & L4        & 373   & 42  & 92   & 507   \\
			% & L5        & 120   & 12  & 33   & 165   \\
			% & L6        & 188   & 37  & 54   & 279   \\
			& Total     & 1176  & 166 & 332  & 1674  \\ \hline
			ACDC                  & Total     & 140    & 20  & 40   & 200   \\ \bottomrule
		\end{tabular}
	}
	
	\label{tab:dataset_distribution}
\end{table}

\begin{figure*}[t]
	\centerline{\includegraphics[width=6.4in]{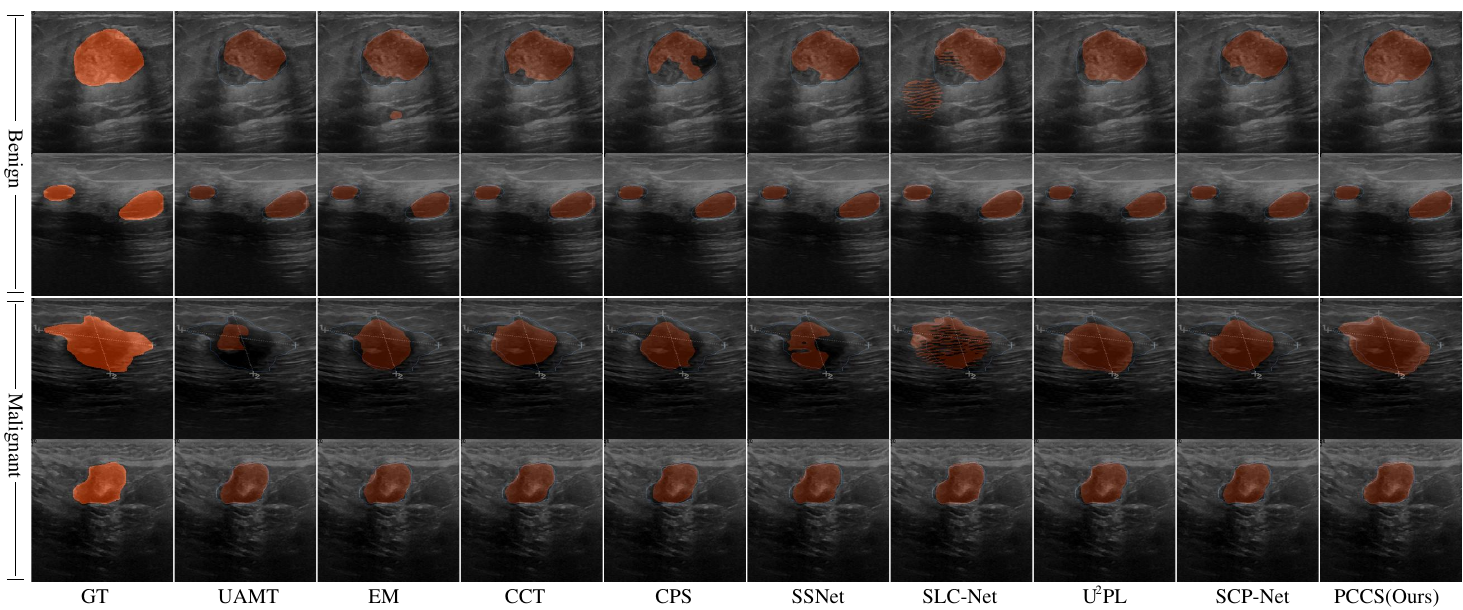}}
	\caption{Visual comparison of segmentation results using different methods on the BUSI dataset.}
	\label{fig:pred_show_busi}
\end{figure*}

\section{Experiments}
\subsection{Setup}
{\raggedleft \bf Datasets.}
We validate the proposed method on several available datasets, the detailed information is described as follows.

BUSI dataset \cite{BUSI} is a dataset consisting of 647 breast ultrasound images,  which is categorized into two classes: benign, and malignant. We split it into a training set, a validation set and a test set, in which the training set includes 452 images, the validation set includes 64 images and the test set includes 131 images.

The Breast MRI dataset (BML) is a new dataset collected by us that consists of 1647 breast MRI images. Multiple modalities including MIP, T1a, and T2 are provided for algorithm research. The dataset is divided in a ratio of 7:1:2, 1176 images form the train set, 166 images form the validation set, and 332 images form the test set.

ACDC dataset \cite{dataset-ACDC} contains 200 annotated short-axis cardiac MR frames from 100 patients. The segmentation masks of the left ventricle (LV), myocardium (Myo), and right ventricle (RV) are provided for clinical and algorithm research. We divide the dataset in the ratio of 7:1:2 to form training set (140 frames), validation set (20 frames), and test set (40 frames).

{\raggedleft \bf Preprocessing and Metrics.} Various data augmentation techniques are employed, including random flipping, random cropping, and adding Gaussian noise to images. Additionally, image resizing to {$224 \times 224$} and grayscale normalization to zero mean and unit variance are performed. The same preprocessing steps are applied to the BUSI, BML, and ACDC datasets. The train set is further divided into a small portion of labeled data, with the remainder serving as unlabeled data. For 2D breast tumor segmentation in the BUSI and BML datasets, the Dice coefficient (referred to as Dice) and the Jaccard coefficient (referred to as Jaccard) were selected as performance metrics. For 3D cardiac segmentation in the ACDC dataset, in addition to Dice and Jaccard, the 95th percentile Hausdorff distance (referred to as 95HD) and the average symmetric surface distance (referred to as ASSD) were utilized as evaluation metrics for 3D image segmentation performance.

{\raggedleft \bf Implementation Details.} The method is implemented on PyTorch with NVIDIA GeForce RTX 4090 GPU. We adopt a 2D architecture for segmenting 2D breast tumor images and cardiac slices. Specifically, the widely used 2D U-Net encoder served as the backbone. The input image size is set to {$224 \times 224$} pixels, and the batch size is set to 16, including 8 labeled images and 8 unlabeled images. The number of channels of the feature map used for contrastive sample sampling in the encoder is set to 256. The maximum training step, denoted as {$t_{max}$}, is set to 20000. We trained the network using the SGD optimizer (weight decay = 0.0001, momentum = 0.9). The learning rate is initialized as 0.05 and decayed by multiplication with {$(1.0 - t/t_{max})^{0.9}$}.

\subsection{Baseline Approaches} 
We compare our method with the supervised method (UNet) \cite{UNet} and state-of-the-art semi-supervised medical image segmentation methods, including Mean-Teacher self-ensembling model (MT) \cite{Consis-MT}, Entropy Minimization approach  (EM) \cite{EM}, Uncertainty-Aware MT (UA-MT) \cite{Consis-Uncertainty-UA-MT}, Cross-Consistency Training (CCT) \cite{Consis-CCT}, Cross Pseudo Supervision (CPS) \cite{pseudo-CPS}, Mutual Consistency-Network (MC-Net) \cite{MCNet-consistency}, SS-Net \cite{SSNet-consistency}, SLC-Net \cite{SLC-Net}, U{$^2$}PL \cite{pseudo-contra-U2PL}, SCP-Net \cite{SCPNet-consistency}. All the methods are implemented with the same backbone and training protocols to ensure fairness.

\begin{table}[htbp]
	\centering
	
	% \tiny
	% \scriptsize
	% \footnotesize
	\small
	\caption{Comparison of different methods on the BUSI dataset under semi-supervised setting. PCCS is SOTA on average. L represents the labeled image and U represents the unlabled image.}
	\setlength{\tabcolsep}{0.3mm}{
		\begin{tabular}{c|c|cc|cc|cc}
			\hline
			\multirow{2}{*}{\textbf{Method}} & \textbf{\#Scans Used}       & \multicolumn{2}{c|}{\textbf{Benign}} & \multicolumn{2}{c|}{\textbf{Malignant}} & \multicolumn{2}{c}{\textbf{Average}} \\ \cline{2-8} 
			& L,U                         & Dice(\%)          & Jaccard(\%)      & Dice(\%)           & Jaccard(\%)        & Dice(\%)          & Jaccard(\%)      \\ \hline
			UNet                             & 10\%,0                      & 56.84             & 48.81            & 22.33              & 17.76              & 45.78             & 38.85            \\ \hline
			MT                               & \multirow{11}{*}{10\%,90\%} & 49.47             & 41.38            & 34.24              & 26.21              & 44.59             & 36.51            \\
			UAMT                             &                             & \textbf{61.30}    & \textbf{52.71}   & 30.82              & 23.77              & 51.53             & 43.43            \\
			CCT                              &                             & 56.63             & 47.06            & 31.68              & 23.77              & 48.63             & 39.6             \\
			EM                               &                             & 57.98             & 49.52            & 21.5               & 16.71              & 46.28             & 39               \\
			CPS                              &                             & 56.21             & 48.29            & 27.7               & 21.47              & 47.07             & 39.69            \\
			MC-Net                           &                             & 57.29             & 48.38            & 25.24              & 19.01              & 47.02             & 38.96            \\
			SSNet                            &                             & 60.00             & 51.21            & 33.01              & 25.75              & 51.35             & 43.05            \\
			SLC-Net                          &                             & 57.83             & 48.49            & 38.9               & 30.14              & 51.76             & 42.61            \\
			U{$^2$}PL                             &                             & 53.19             & 45.83            & 20.56              & 15.6               & 42.73             & 36.14            \\
			SCP-Net                          &                             & 58.65             & 50.37            & 36.09              & 27.97              & 51.42             & 43.19            \\
			PCCS                             &                             & 57.44             & 48.38            & \textbf{47.67}     & \textbf{35.75}     & \textbf{54.31}    & \textbf{44.33}   \\ \hline
			UNet                             & 20\%,0                      & 63.97             & 56.9             & 27.11              & 21.63              & 52.15             & 45.59            \\ \hline
			MT                               & \multirow{11}{*}{20\%,80\%} & 68.70             & 59.11            & 53.26              & 42.65              & 63.75             & 53.83            \\
			UAMT                             &                             & 70.39             & 62.31            & 43.33              & 33.99              & 61.71             & 53.23            \\
			CCT                              &                             & 67.14             & 57.65            & 45.39              & 35.57              & 60.17             & 50.57            \\
			EM                               &                             & 65.64             & 56.21            & 42.89              & 32.48              & 58.34             & 48.61            \\
			CPS                              &                             & 65.66             & 56.07            & 50.61              & 39.75              & 60.84             & 50.84            \\
			MC-Net                           &                             & 64.65             & 55.8             & 40.43              & 31.95              & 56.89             & 48.15            \\
			SSNet                            &                             & \textbf{71.35}    & \textbf{63.07}   & 41.05              & 32.50              & 61.64             & 53.27            \\
			SLC-Net                          &                             & 65.08             & 54.86            & 52.31              & 39.96              & 60.98             & 50.08            \\
			U{$^2$}PL                               &                             & 68.57             & 60.1             & 49.91              & 39.39              & 62.58             & 53.46            \\
			SCP-Net                          &                             & 70.35             & 62.77            & 48.94              & 38.32              & 63.49             & 54.93            \\
			PCCS                             &                             & 68.04             & 58.65            & \textbf{63.31}     & \textbf{48.93}     & \textbf{66.52}    & \textbf{55.53}   \\ \hline
			UNet                             & 100\%,0                     & 70.09             & 59.14            & 74.91              & 63.21              & 71.63             & 60.45            \\ \hline
		\end{tabular}
	}
	
	\label{tab:main_result_on_BUSI}
\end{table}

\begin{figure*}[t]
	\centerline{\includegraphics[width=6.4in]{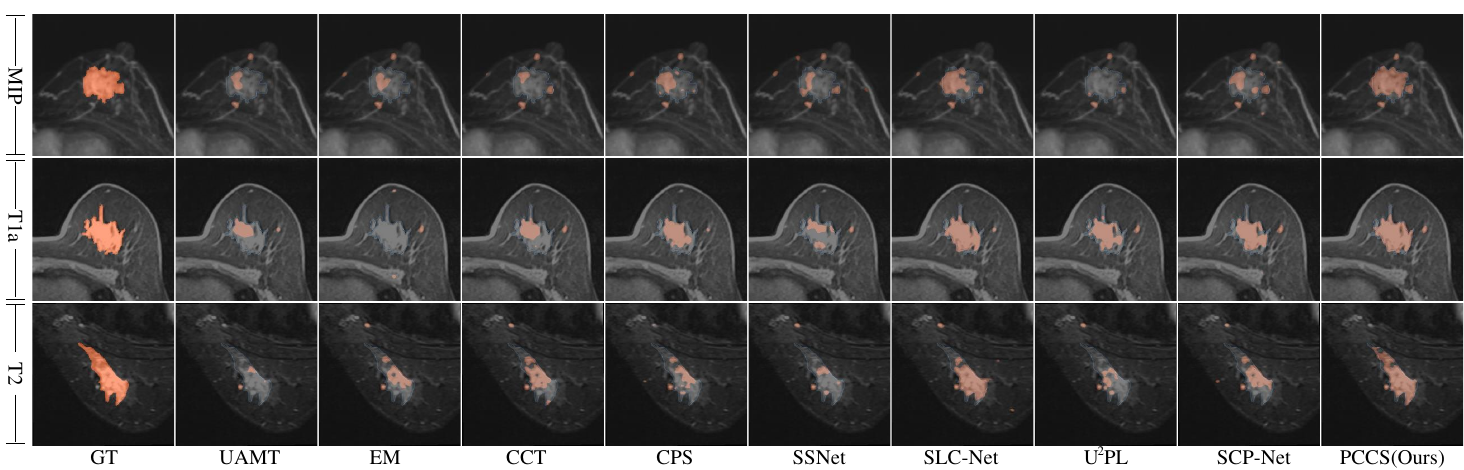}}
	\caption{Visual comparison of segmentation results using different methods on the BML dataset.}
	\label{fig:pred_show_BML}
\end{figure*}

\subsection{Quantitative Comparison of Breast Tumor Images}
In order to comprehensively evaluate the effectiveness of PCCS, we will conduct experiments on two modalities (ultrasound image and magnetic resonance image) of breast tumor.
\subsubsection{Comparison results on BUSI}
{\raggedleft \bf Results of Comparison Methods.} Table~\ref{tab:main_result_on_BUSI} reports the experimental results of PCCS on BUSI under two semi-supervised settings. PCCS significantly outperforms other state-of-the-art (SOTA) methods on average. Specifically, malignant tumors often present irregular lesion boundaries and low contrast between lesion regions and background, posing challenges for accurate segmentation. When using 20\% of annotated images, the best-performing SOTA method achieved the highest segmentation performance on malignant tumors (Dice=53.26\%, Jaccard=42.65\%), while PCCS achieved segmentation performance on malignant tumors (Dice=63.31\%, Jaccard=48.93\%). PCCS outperforms the best-performing SOTA method by 10.05\% and 6.28\% in terms of Dice and Jaccard indices, respectively. Even with a reduced number of annotated samples in the training set (only 10\% of annotated images), PCCS remains highly competitive and significantly outperforms other methods.

{\raggedleft \bf Results of Comparison on Visualization.} Fig.~\ref{fig:pred_show_busi} shows the visualization results of various methods applied to the BUSI dataset when using 20\% annotated images and 80\% unannotated images. A comparative analysis is conducted among nine methods: UAMT, EM, CCT, CPS, SSNet, SCL-Net, U{$^2$}PL, SCP-Net, and PCCS. The leftmost column represents the Ground Truth (denoted as GT), providing a reference for comparison. The first row showcases the segmentation outcomes of benign tumors, while the second row demonstrates the segmentation results of malignant tumor. It is noticeable that benign tumor in breast ultrasound image typically displays relatively regular boundaries resembling ellipses, whereas malignant tumor in breast ultrasound image tend to exhibit irregular boundaries with sharp protrusions. Traditional semi-supervised methods often excel in learning features associated with benign tumors characterized by regular boundaries, yet struggle to accurately segment malignant tumors with irregular boundaries. In contrast, PCCS effectively constructs relationships among boundary prototypes, enabling the learning of more discriminative features. Consequently, the segmentation results produced by PCCS for malignant tumors demonstrate a closer alignment with the true boundaries of the Ground Truth, exhibiting fewer missing and unexpected regions.

\begin{table}[!htbp]
	\centering
	%% \tiny
	% \scriptsize
	% \footnotesize
	\small
	% \normalsize
	% \large
	\caption{Comparison of different methods on the BML datasets under semi-supervised setting. PCCS is SOTA on average. L represents the labeled image and U represents the unlabled image.}
	\setlength{\tabcolsep}{0.2mm}{
		\begin{tabular}{c|c|cc|cc|cc|cc}
			\hline
			\multirow{2}{*}{\textbf{Method}} & \textbf{\#Scans Used}       & \multicolumn{2}{c|}{\textbf{MIP}} & \multicolumn{2}{c|}{\textbf{T1a}} & \multicolumn{2}{c|}{\textbf{T2}} & \multicolumn{2}{c}{\textbf{Average}} \\ \cline{2-10} 
			& L,U                         & Dice(\%)        & Jaccard(\%)     & Dice(\%)        & Jaccard(\%)     & Dice(\%)        & Jaccard(\%)    & Dice(\%)          & Jaccard(\%)      \\ \hline
			UNet                             & 10\%,0                      & 35.25           & 25.7            & 39.63           & 28.86           & 33.13           & 22.92          & 36.08             & 25.87            \\ \hline
			MT                               & \multirow{11}{*}{10\%,90\%} & 35.43           & 25.97           & 41.88           & 31.41           & 34.73           & 24.59          & 37.48             & 27.43            \\
			UAMT                             &                             & 43.83           & 31.3            & 45.44           & 32.81           & 38.39           & 26.54          & 42.53             & 30.2             \\
			CCT                              &                             & 38.88           & 27.17           & 43.98           & 31.46           & 37.54           & 25.83          & 40.23             & 28.23            \\
			EM                               &                             & 38.52           & 28.42           & 43.71           & 32.75           & 38.97           & 27.58          & 40.52             & 29.67            \\
			CPS                              &                             & 34.81           & 25.34           & 41.84           & 31.61           & 34.51           & 25.28          & 37.21             & 27.55            \\
			MC-Net                           &                             & 40.29           & 29.18           & 46.84           & 35.17           & 37.76           & 26.61          & 41.75             & 30.43            \\
			SSNet                            &                             & 42.19           & 30.06           & 49.1            & 36.36           & 39.39           & 27.38          & 43.68             & 31.38            \\
			SLC-Net                          &                             & 44.83           & 31.79           & 44.81           & 31.01           & 40.17           & 28.3           & 43.22             & 30.32            \\
			U{$^2$}PL                             &                             & 42.83           & 33.05           & 47.46           & 36.26           & 45              & 33.73          & 45.22             & 34.42            \\
			SCP-Net                          &                             & 34.5            & 25.84           & 41.84           & 32.56           & 36.6            & 26.79          & 37.83             & 28.55            \\
			PCCS                             &                             & \textbf{57.84}  & \textbf{46.82}  & \textbf{62.88}  & \textbf{51.68}  & \textbf{53.33}  & \textbf{41.64} & \textbf{58.08}    & \textbf{46.77}   \\ \hline
			UNet                             & 20\%,0                      & 47.87           & 35.8            & 53.89           & 41.82           & 42.06           & 30.43          & 48.02             & 36.1             \\ \hline
			MT                               & \multirow{11}{*}{20\%,80\%} & 51.42           & 38.27           & 52.78           & 40.37           & 41.81           & 30.02          & 48.61             & 36.18            \\
			UAMT                             &                             & 49.77           & 39.06           & 55.79           & 45.1            & 42.63           & 32.19          & 49.46             & 38.85            \\
			CCT                              &                             & 49.43           & 38.22           & 54.99           & 42.76           & 40.54           & 29.42          & 48.35             & 36.81            \\
			EM                               &                             & 46.45           & 36.08           & 54.58           & 43.74           & 43.75           & 33             & 48.41             & 37.75            \\
			CPS                              &                             & 48.14           & 36.39           & 55.78           & 43.86           & 43.28           & 31.62          & 49.19             & 37.4             \\
			MC-Net                           &                             & 51.46           & 39.75           & 59.74           & 47.69           & 44.24           & 33.21          & 51.92             & 40.33            \\
			SSNet                            &                             & 50.34           & 38.82           & 56.71           & 45.64           & 41.48           & 31.28          & 49.56             & 38.65            \\
			SLC-Net                          &                             & 55.65           & 43.7            & 56.32           & 45.37           & 46.68           & 35.64          & 52.81             & 41.52            \\
			U{$^2$}PL                             &                             & 49.79           & 38.79           & 55.48           & 44.07           & 43.38           & 32.63          & 49.61             & 38.55            \\
			SCP-Net                          &                             & 51.66           & 39.5            & 57.4            & 44.88           & 43.87           & 32.57          & 51.03             & 39.03            \\
			PCCS                             &                             & \textbf{62.79}  & \textbf{52.03}  & \textbf{66.69}  & \textbf{56.43}  & \textbf{56.64}  & \textbf{45.73} & \textbf{62.06}    & \textbf{51.43}   \\ \hline
			UNet                             & 100\%,0                     & 69.68           & 58.69           & 67.75           & 57.76           & 58.35           & 47.39          & 65.10             & 54.48            \\ \hline
		\end{tabular}
	}
	
	\label{tab:main_result_on_BML}
\end{table}

\subsubsection{Comparison results on BML}
Table~\ref{tab:main_result_on_BML} presents the experimental findings on BML. Notably, PCCS demonstrates superior performance compared to state-of-the-art methods across MIP, T1a, and T2 modalities. Specifically, when leveraging 20\% annotated images and 80\% unannotated images, the state-of-the-art approach (SLC-Net) achieves an average segmentation performance with a Dice score of 52.81\% and a Jaccard index of 41.52\%, whereas PCCS achieves notably higher scores with a Dice score of 62.06\% and a Jaccard index of 51.43\%. These results signify a significant improvement of 10.85\% and 9.91\% in Dice and Jaccard scores, respectively, compared to SLC-Net. Remarkably, even with a reduction in the number of annotated samples within the training set to only 10\% annotated images and 90\% unannotated images, PCCS remains highly competitive and outperforms other methods significantly.

{\raggedleft \bf Results of Comparison on Visualization.} Fig.~\ref{fig:pred_show_busi} showcases the visualization outcomes of various methods applied to the BUSI dataset when employing 20\% annotated images and 80\% unannotated images. A comprehensive comparative analysis is conducted among nine methods: UAMT, EM, CCT, CPS, SSNet, SCL-Net, U{$^2$}PL, SCP-Net, and PCCS. The leftmost column illustrates the Ground Truth (referred to as GT), serving as a reference for comparison. Subsequently, the first, second, and third rows present the segmentation visual results for MIP, T1a, and T2, respectively. Observing the visualizations, it becomes apparent that the boundaries of lesion areas exhibit irregular shapes, and there are also some spot-like signals present in non-lesion areas. These irregular shapes and spot-like noise present challenges for precise segmentation. While existing semi-supervised methods excel at learning features with regular shapes within lesion areas, they often struggle to accurately segment areas with irregular boundaries. In contrast, PCCS effectively establishes relationships between uncertainty-guided boundary prototypes, facilitating the learning of more discriminative features and mitigating noise caused by spot-like signals. Consequently, the segmentation results produced by PCCS across the three modalities demonstrate greater consistency with the true boundaries of the ground truth, exhibiting fewer missing and unexpected regions.

\begin{figure}[htbp]
	\centerline{\includegraphics[width=2.8in]{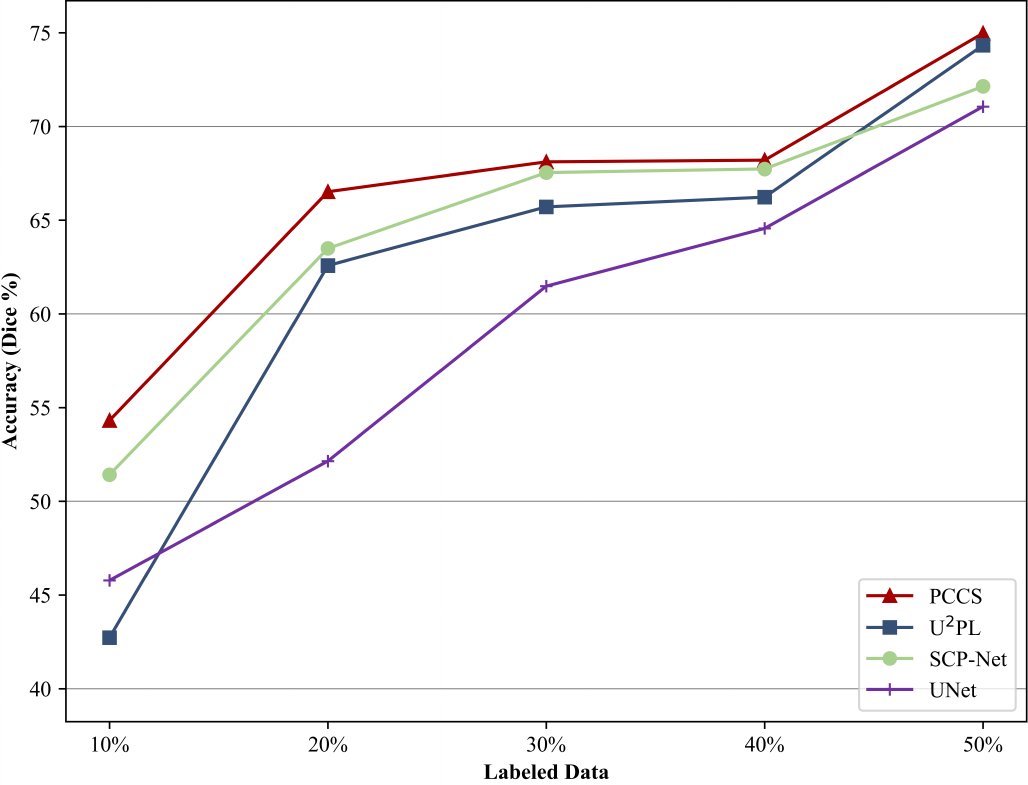}}
	\caption{Quantitative results on different semi-supervised settings on BUSI.}
	\label{fig:multi-semi_on_busi}
\end{figure}

\subsection{Ablation Studies}

{\raggedleft \bf Effectiveness of Each Component.}
Table~\ref{tab:ablation_BUSI} presents the ablation study results of the proposed PCCS on the BUSI dataset, comprising two sets of semi-supervised settings: 10\% labeled and 90\% unlabeled images (referred to as 10\%) and 20\% labeled and 80\% unlabeled images (referred to as 20\%). Taking the case of 20\%, by employing all components listed in Table~\ref{tab:ablation_BUSI}, PCCS achieves the optimal performance (Dice=66.52\%, Jaccard=55.53\%). The accuracy of Dice and Jaccard indicates that each component contributes positively to PCCS, implying a continual improvement in model performance. In scenarios with fewer labeled data (10\%), each component of PCCS still makes a positive contribution, continuously enhancing the model's performance. These results demonstrate the effectiveness of the proposed method.

\begin{table}[t]
	\centering
	\begin{minipage}{0.46\linewidth} % 调整宽度比例
		\centering
		%\tiny
		% \scriptsize
		% \footnotesize
		%\small
		\normalsize
		% \large
		\caption{Quantitative results of ablation study of PCCS.}
		\resizebox{0.93\linewidth}{!}{
		\begin{tabular}{cccccccc}
				\toprule
				\multirow{2}{*}{{$l_{con}$}} & \multirow{2}{*}{{$l_{u}$}} & \multirow{2}{*}{{$l_{aux}$}} & \multirow{2}{*}{{$l_{pc}$}} & \multicolumn{2}{c}{10\%}        & \multicolumn{2}{c}{20\%}        \\ \cline{5-8} 
				&                        &                       &                        & Dice(\%)           & Jaccard(\%)        & Dice(\%)           & Jaccard(\%)        \\ \hline
				\checkmark                       &                        &                       &                        & 53.77          & 45.14          & 58.31          & 50.08          \\
				& \checkmark                      &                       &                        & 51.95          & 41.49          & 59.24          & 51.03          \\
				\checkmark                       & \checkmark                      &                       &                        & 47.68          & 40.30          & 59.47          & 51.28          \\
				&                        & \checkmark                     & \checkmark                      & 46.37          & 38.67          & 61.34          & 52.08          \\
				\checkmark                       & \checkmark                      & \checkmark                     &                        & 50.72          & 42.69          & 61.52          & 52.62          \\
				\checkmark                       & \checkmark                      & \checkmark                     & \checkmark                      & \textbf{54.31} & \textbf{44.33} & \textbf{66.52} & \textbf{55.53} \\ \bottomrule
			\end{tabular}
		}
		\label{tab:ablation_BUSI}
	\end{minipage}
	\hfill % 添加水平间距
	\begin{minipage}{0.46\linewidth} % 调整宽度比例
		\centering
		%\tiny
		% \scriptsize
		% \footnotesize
		%\small
		 \normalsize
		% \large
		\caption{The results (Dice(\%)) of different coefficients.}
		\resizebox{0.7\linewidth}{!}{
			\begin{tabular}{cccc|ccc}
				\toprule
				\multirow{2}{*}{Value} & \multicolumn{3}{c|}{10\%}                     & \multicolumn{3}{c}{20\%}                         \\ \cline{2-7} 
				& {$\lambda_{pc}$}    & {$\lambda_{u}$}    & {$\lambda_{c}$}           & {$\lambda_{pc}$}    & {$\lambda_{u}$}    & {$\lambda_{c}$}              \\ \hline
				{$10^{-4}$}                 & 51.8           & 50.46          & 48.92       & 57.91          & 58.16          & 59.52          \\
				{$10^{-3}$}                  & 52.58          & 51.55          & 54.3        & 62.36          & \textbf{60.55} & 61.62          \\
				{$10^{-2}$}                   & 53.55          & \textbf{54.26} & 51.42       & 59.26          & 59.58          & 59.68          \\
				{$10^{-1}$}                    & 52.19          & 52.14          & \textbf{54} & 59.36          & 56.73          & \textbf{66.52} \\
				{$10^0$}                     & \textbf{54.31} & 45.86          & 42.7        & \textbf{63.19} & 51.11          & 56.03          \\
				{$10^{1}$}                     & 47.25          & 29.99          & 36.29       & 56.8           & 31.08          & 47.91          \\ \bottomrule
			\end{tabular}
		}
		\label{tab:contra_co_busi}
	\end{minipage}
\end{table}

\begin{table*}[htbp]
	\centering
	% \tiny
	% \scriptsize
	\footnotesize
	% \small
	% \normalsize
	% \large
	\caption{Comparison of different methods on the ACDC dataset under semi-supervised setting. Let L denote the labeled image and U denote the unlabled image. PCCS is SOTA on average.}
	\setlength{\tabcolsep}{0.8mm}{
		\begin{tabular}{c|c|cc|cc|cc|cc}
			\specialrule{.1em}{0em}{0em}
			\multirow{2}{*}{\textbf{Method}} & \textbf{\# Scans Used} & \multicolumn{2}{c|}{\textbf{RV}} & \multicolumn{2}{c|}{\textbf{Myo}} & \multicolumn{2}{c|}{\textbf{LV}} & \multicolumn{2}{c}{\textbf{Average}} \\ \cline{2-10} 
			& L,U       & Dice (\%)            & 95HD (voxel)          & Dice (\%)            & 95HD(voxel)           & Dice (\%)            & 95HD (voxel)          & Dice (\%)              & 95HD (voxel)             \\ \hline
			UNet                             & 14(10\%),0             & 81.06           & 9.20           & 80.46            & 8.38           & 87.45           & 13.34          & 82.99             & 10.31            \\
			UNet                             & 140(100\%),0           & 91.09           & 1.40           & 89.12            & 1.04           & 94.51           & 1.33           & 91.53             & 1.26             \\ \hdashline
			MT                               & 14,126                 & 82.49           & 5.08           & 82.34            & \textbf{4.91}  & 88.38           & 11.18          & 84.41             & 7.06             \\
			UAMT                             & 14,126                 & 82.90           & 10.08          & 82.07            & 9.40           & 87.13           & 16.74          & 84.04             & 12.07            \\
			CCT                              & 14,126                 & 82.29           & 9.26           & 82.04            & 6.43           & \underline{ 89.95}     & \textbf{6.72}  & 84.76             & 7.47             \\
			EM                               & 14,126                 & 82.54           & 4.13           & \underline{ 83.03}      & 6.07           & 89.11           & 10.00          & 84.89             & 6.73             \\
			CPS                              & 14,126                 & 83.24           & 5.03           & 82.12            & 6.41           & 89.33           & 9.08           & 84.90             & 6.84             \\
			MC-Net                           & 14,126                 & 83.68           & 4.17           & 81.81            & 6.02           & 86.34           & 13.85          & 83.94             & 8.01             \\
			SSNet                            & 14,126                 & 83.63           & 6.94           & 82.70            & 7.59           & 88.62           & 10.16          & 84.99             & 8.23             \\
			U{$^2$}PL                             & 14,126                 & \textbf{86.07}  & \textbf{1.92}  & 82.71            & 6.07           & 88.83           & 10.65          & \underline{ 85.87}       & \underline{ 6.22}       \\
			SCP-Net                          & 14,126                 & 82.42           & \underline{ 2.89}     & 81.82            & 6.17           & 88.63           & 10.29          & 84.56             & 6.45             \\
			PCCS                             & 14,126                 & \underline{ 85.07}     & 3.94           & \textbf{84.03}   & \underline{ 4.98}     & \textbf{90.04}  & \underline{ 7.78}     & \textbf{86.38}    & \textbf{5.57}    \\ \bottomrule
		\end{tabular}
		
	}
	
	\label{tab:main_table_on_ACDC}
\end{table*}

{\raggedleft \bf Impact of Different Coefficient in Loss.} Different hyperparameters in {$\lambda_{c}$}, {$\lambda_{u}$}, {$\lambda_{pc}$} have different impacts on PCCS's performances. In order to find a set of suitable coefficients in loss, we test the performance of PCCS on the BUSI dataset with different coefficients. The parameters {$\lambda_{c}$}, {$\lambda_{u}$}, {$\lambda_{aux}$}, {$\lambda_{pc}$} are selected from {$\{10^{-4},10^{-3},...,10\}$}. The loss function's hyperparameters are initialized as {$\lambda_{c}=0.1$}, {$\lambda_{aux}=0.3$}, and {$\lambda_{pc}=0.01$}. Experiments are conducted based on this hyperparameter set. Table~\ref{tab:contra_co_busi} reports the influence of different hyperparameter values under two semi-supervised settings on the performance of PCCS. Taking {$\lambda_{pc}$} as an example, with {$\lambda_{c}=0.1$} and {$\lambda_{aux}=0.3$}, changing {$\lambda_{pc}$} under the 10\% semi-supervised setting results in the Dice value reaching its optimum when {$\lambda_{pc}=1$}. Similarly, under the 20\% semi-supervised setting, PCCS achieves the best segmentation results when {$\lambda_{pc}=1$}. Experimental results from both semi-supervised settings indicate that {$\lambda_{pc}=1$} is optimal, possibly because the model indeed learns useful knowledge during the training phase. If {$\lambda_{pc}$} is set too small, the discriminative ability of the learned prototypes decreases as the model focuses more on other regions. Similarly, considering {$\lambda_{c}$}, Table~\ref{tab:contra_co_busi} demonstrates that {$\lambda_{c}=10^{-1}$} achieves optimal performance under both semi-supervised settings. Specifically, {$\lambda_{u}$} equals {$10^{-2}$} achieves optimum in the 10\% setting, while {$\lambda_{u}$} equals {$10^{-3}$} achieves optimum in the 20\% semi-supervised setting. The uncertainty coefficient should not be set too large, and its value ranges between {$10^{-3}$} and {$10^{-2}$}. In conclusion, the selection of hyperparameters significantly impacts the performance of PCCS on the BUSI dataset. Optimizing $\lambda_{pc}$, $\lambda_{c}$ and $\lambda_{u}$ within appropriate ranges ensures the model's efficacy in capturing relevant features and discriminating between classes under varying semi-supervised scenarios.

\begin{figure}[htbp]
	\centerline{\includegraphics[width=4in]{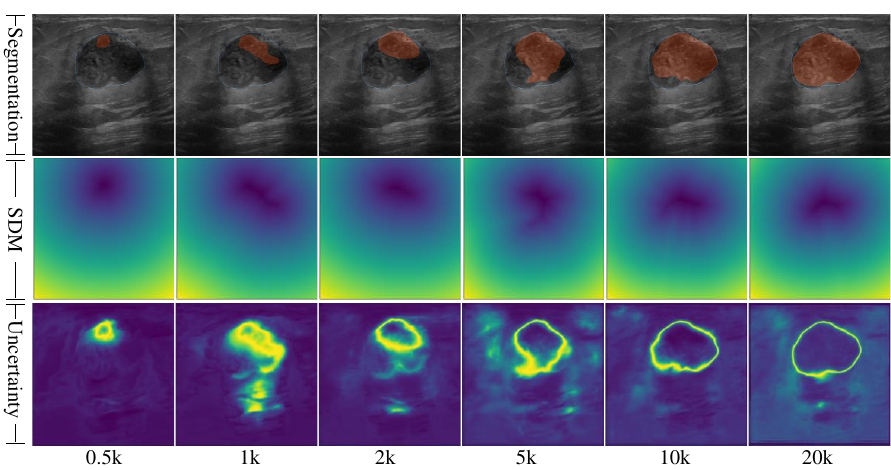}}
	\caption{Visual comparison of segmentation map, signed distance map (denoted SDM), and uncertainty map in different training stages on BUSI dataset.}
	\label{fig:uncertainty_sdm_on_busi}
\end{figure}

{\raggedleft \bf Visualization of Segmentation Map, Singed Distance Map, and Uncertainty Map.} We visualize segmentation map, signed distance map and uncertainty map at various training stages. As shown in Fig.~\ref{fig:uncertainty_sdm_on_busi}, the segmentation results gradually cover the entire lesion area, as observed in the first row. The signed distance map also varies with the segmentation results, as seen in the second row. Due to a significant reduction in prediction uncertainty, PCCS can gradually reduce the uncertainty region and eventually compresses the uncertainty region around the segmentation boundary, as depicted in the third row.

{\raggedleft \bf Performance of Different semi-supervised settings.}  Fig.~\ref{fig:multi-semi_on_busi} shows the average performance of PCCS, U{$^2$}PL, SCP-Net, and UNet under different semi-supervised settings (with labeled images in the training set accounting for 10\%, 20\%, 30\%, 40\%, and 50\%, respectively) on BUSI. It is evident that as the proportion of labeled data increases, the segmentation accuracy (Dice) of each method also improves. When labeled images constitute 50\% of the dataset, the Dice scores of all methods exceed 70. Conversely, as the proportion of labeled data decreases, the segmentation accuracy of each method also decreases. Taking PCCS as an example, the Dice scores corresponding to 50\%, 40\%, 30\%, 20\%, and 10\% labeled data are 74.97\%, 68.20\%, 68.11\%, 66.52\%, and 54.31\%, respectively. From 50\% to 20\%, the Dice score decreases from 74.97\% to 66.52\%, with an average decrease of 2.11\%. Further reduction from 20\% to 10\% results in a decrease of 12.21\% in Dice score, indicating a significant decline in segmentation accuracy. Therefore, we conclude that labeled images continue to have a substantial impact on the performance of semi-supervised methods.

\begin{figure*}[t]
	\centerline{\includegraphics[width=6.2in]{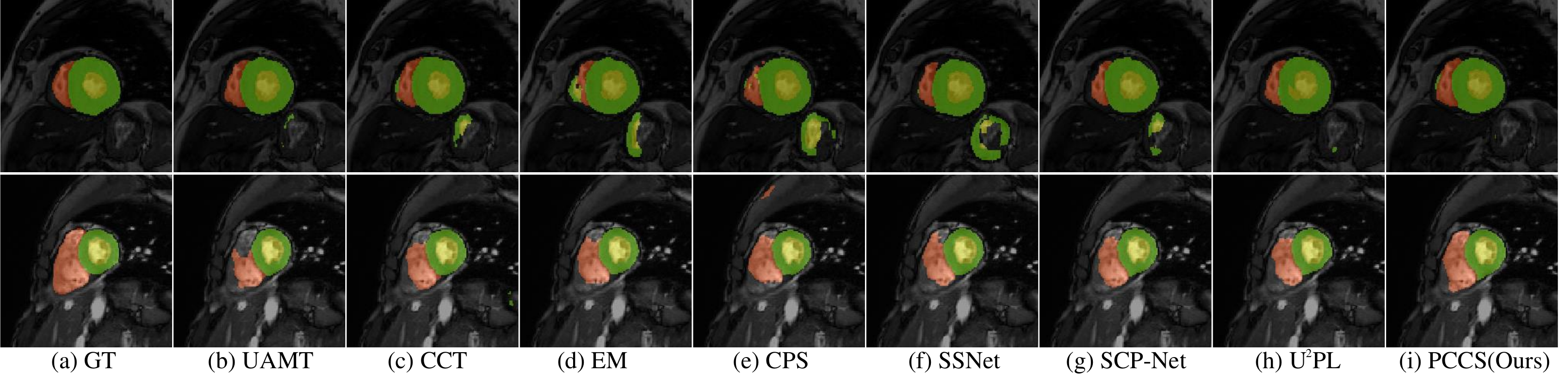}}
	\caption{Visual comparison of segmentation results using different methods on the ACDC dataset. GT represents Ground Truth. PCCS can perform accurate segmentation, while other methods have omissions and wrong segmentation.}
	\label{fig:pred_show_acdc}
\end{figure*}

\begin{table}[htbp]
	\centering
	% \tiny
	% \scriptsize
	% \footnotesize
	\small
	% \normalsize
	% \large
	\caption{The effect of different encoder layers. (v) represents (voxel). Fuse middle represents a feature that fuses Conv2, Conv3, and Conv4, its resolution is {$56 \times 56$}. Fuse All represents a feature that fuses Conv1, Conv2, Conv3, Conv4, and Conv5, its resolution is {$56 \times 56$}.}
	\setlength{\tabcolsep}{0.3mm}{
		\begin{tabular}{c|cc|cc|cc|cc}
			\hline
			\multirow{2}{*}{\begin{tabular}[c]{@{}c@{}}Feature\\ (Resolution)\end{tabular}} & \multicolumn{2}{c|}{RV}        & \multicolumn{2}{c|}{Myo}       & \multicolumn{2}{c|}{LV}        & \multicolumn{2}{c}{Average}    \\ \cline{2-9} 
			& Dice(\%)       & 95HD(v)    & Dice(\%)       & 95HD(v)   & Dice(\%)       & 95HD(v)   & Dice(\%)       & 95HD(v)    \\ \hline
			\begin{tabular}[c]{@{}c@{}}Conv1 \\ ({$224 \times 224$})\end{tabular}                      & 80.17          & 2.75          & 80.8           & 5.18          & 86.47          & 10.65         & 82.48          & 6.2           \\ \hline
			\begin{tabular}[c]{@{}c@{}}Conv2 \\ ({$112 \times 112$})\end{tabular}                      & 80.91          & 2.78          & 82.01          & 3.27          & 88.15          & 11.69         & 83.69          & 5.91          \\ \cline{1-8}
			\begin{tabular}[c]{@{}c@{}}Conv3 \\ ({$56 \times 56$})\end{tabular}                        & 82.27          & 2.08          & 81.41          & 5.88          & 88.8           & 6.89          & 84.16          & 4.95          \\ \hline
			\begin{tabular}[c]{@{}c@{}}Conv4 \\ ({$28 \times 28$})\end{tabular}                        & 82.37          & 2.2           & 80.71          & 4.31          & 87.31          & 6.53          & 83.46          & 4.35          \\ \hline
			\begin{tabular}[c]{@{}c@{}}Conv5 \\ ({$14 \times 14$})\end{tabular}                        & 80.97          & 8.05          & 80.34          & 5.45          & 84.95          & 14.79         & 82.09          & 9.43          \\ \hline
			\begin{tabular}[c]{@{}c@{}}Fuse Middle \\ ({$56 \times 56$})\end{tabular}                  & \textbf{85.07} & 3.94          & \textbf{84.03} & 4.98          & \textbf{90.04} & 7.78          & \textbf{86.38} & 5.57          \\ \hline
			\begin{tabular}[c]{@{}c@{}}Fuse All \\ ({$56 \times 56$})\end{tabular}                     & 84.21          & \textbf{1.98} & 82.3           & \textbf{2.41} & 88.76          & \textbf{5.25} & 85.09          & \textbf{3.21} \\ \hline
		\end{tabular}
	}
	
	\label{tab:diff_layer_on_acdc}
\end{table}

\subsection{Extended Study of PCCS on Multi-class Segmentation}
In order to further assess the applicability and effectiveness of PCCS, we further extend PCCS to the multi-class segmentation task of cardiac magnetic resonance images.

{\raggedleft \bf Results of Comparison Methods on ACDC.} Table~\ref{tab:main_table_on_ACDC} shows the experimental results of the proposed PCCS on the ACDC when using 10\% labeled data and 90\% unlabeled data. In particular, for Dice, PCCS achieves 85.07\%, 84.03\%, and 90.04\% in RV, Myo, and LV, respectively. PCCS excels in Myo and LV, while being slightly lower in RV compared to U{$^2$}PL (86.07\%). For 95HD, PCCS remains competitive across all classes. In summary, Table~\ref{tab:main_table_on_ACDC} indicates that while other methods perform well only in a single class, PCCS performs well across multiple classes. Therefore, PCCS outperforms other methods in average performance.

{\raggedleft \bf Comparison on Visualization on ACDC.} Fig.~\ref{fig:pred_show_acdc} shows some visualization results on different methods when using 10\% labeled images on the ACDC dataset. We choose four representative methods to show the superior performance of PCCS. Not only is the prediction performance of PCCS significantly higher than SSNet, SCP-Net, U{$^2$}PL, but also the visualization results are closer to the ground truth label with fewer missing and unexpected regions.

{\raggedleft \bf Impact of Different Encoder Layers for Contrastive Learning.} In the proposed PCCS, we select the fusion of intermediate feature maps from the 3 layers of the backbone network as the feature for sampling the prototypes for contrastive learning. Results for different encoder layers are presented in Table~\ref{tab:diff_layer_on_acdc}. Among all individual encoder layers, Conv3 performs the best (Dice=84.16\%). Larger feature maps contain more low-level details due to their higher resolution, while smaller feature maps incorporate more high-level features. We further choose to concatenate features for fusion from multiple encoder layers to obtain more semantic information. It can be seen that the fusion of features from multiple encoder layers demonstrates superior performance, i.e. Fuse middle. Therefore, fusion feature maps in PCCS containing more semantic information lead to better performance.

\begin{figure}[!t]
	\centerline{\includegraphics[width=4in]{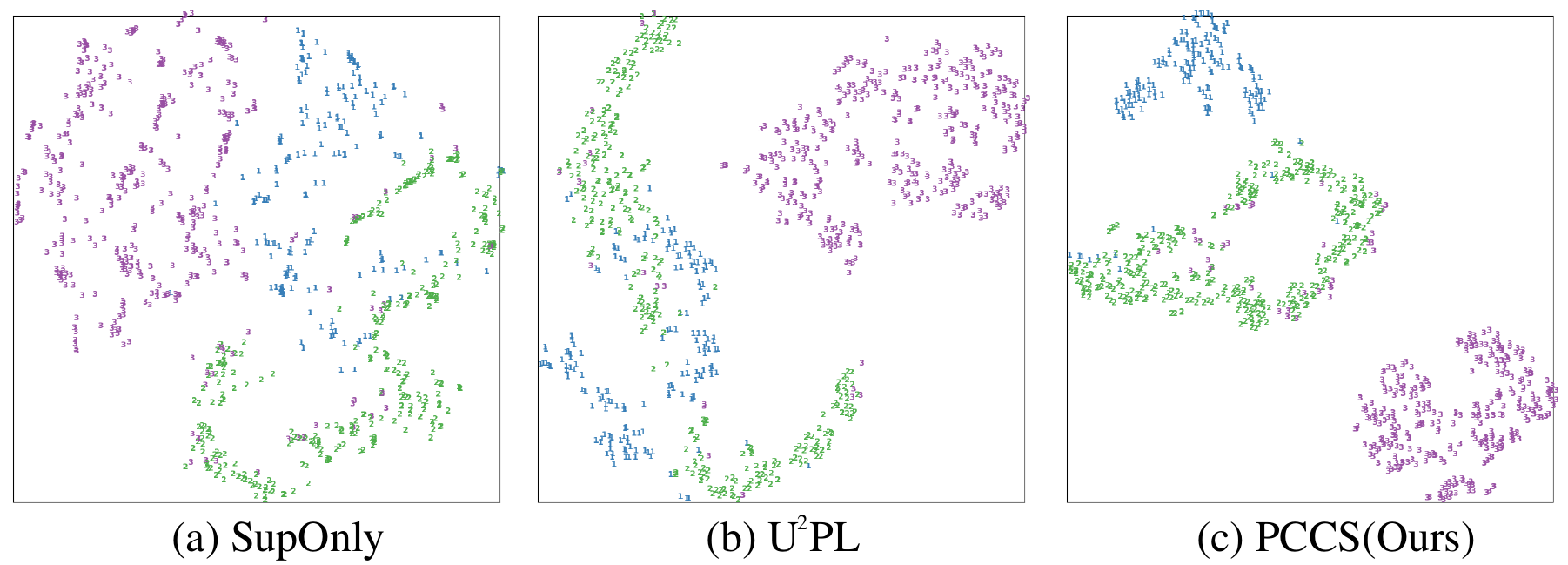}}
	\caption{Visualization results of the pixel features obtained by different methods on ACDC dataset. The dimension of pixel features is reduced by t-SNE algorithm. Colors represent pixel categories. (a) is the SupOnly without contrastive learning, (b) is U{$^2$}PL, and (c) is PCCS.}
	\label{fig:contra_on_acdc}
\end{figure}

{\raggedleft \bf Visualization of Features.} In Fig.~\ref{fig:contra_on_acdc}, we use t-SNE algorithm to visualize the pixel features. The purple dot represents LV, the green dot represents Myo, and the blue dot represents RV. Fig.~\ref{fig:contra_on_acdc}(a) is the SupOnly without contrastive learning. Fig.~\ref{fig:contra_on_acdc}(b) is U{$^2$}PL with contrastive learning. Compared with Fig.~\ref{fig:contra_on_acdc}(a) and Fig.~\ref{fig:contra_on_acdc}(b), the proposed PCCS (see Fig.~\ref{fig:contra_on_acdc}(c)) has better intra-class compactness and inter-class separability, which indicates the effectiveness of contrastive learning for medical image segmentation task.

\section{Conclusion}
Previous contrastive learning methods ignore the overall context information of unlabeled image and do not mine the relationship among the prototypes in segmentation. To address this problem, we propose a novel prototype contrastive consistency learning method for semi-supervised medical image segmentation based on prototype contrastive learning and prototype consistency learning. The new uncertainty-weighted prototype contrastive consistency loss and prototype updating mechanism are designed to promote the effectiveness of prototype contrastive learning. It's found that enhancing the discriminative ability of the prototypes from the feature map can promote the performance of the model. Extensive experiments show that the proposed method outperforms existing methods on several semi-supervised medical image segmentation datasets.

%\bibliographystyle{unsrt}  
%\bibliography{references}  %%% Remove comment to use the external .bib file (using bibtex).

\end{document}